# Title: Toward Scalable Early Cancer Detection: Evaluating EHR-Based Predictive Models Against Traditional Screening Criteria

**Authors:** Jiheum Park[1*], Chao Pang[2], Tristan Y. Lee[3], Jeong Yun Yang[1], Jacob Berkowitz[4,5], Alexander Z. Wei[3], Nicholas Tatonetti[4,5]

**Affiliations:**

[1]Department of Medicine, Columbia University Irving Medical Center; New York, NY, USA

[2]Department of Biomedical Informatics, Columbia University; New York, NY, USA

[3]Division of Hematology & Oncology, Columbia University Irving Medical Center, New York, NY, USA

[4]Department of Computational Biomedicine, Cedars-Sinai Medical Center; Los Angeles, CA, USA.

[5]Cedars-Sinai Cancer, Cedars-Sinai Medical Center; Los Angeles, CA, USA.

*Corresponding author. Email: jp4147@cumc.columbia.edu

**Abstract**

Current cancer screening guidelines cover only a few cancer types and rely on narrowly defined criteria such as age or a single risk factor like smoking history, to identify high-risk individuals. Predictive models using electronic health records (EHRs), which capture large-scale longitudinal patient-level health information, may provide a more effective tool for identifying high-risk groups by detecting subtle prediagnostic signals of cancer. Recent advances in large language and foundation models have further expanded this potential, yet evidence remains limited on how useful EHR-based models are compared with traditional risk factors currently used in screening guidelines. We systematically evaluated the clinical utility of EHR-based predictive models against traditional risk factors, including gene mutations and family history of cancer, for identifying high-risk individuals across eight major cancers (breast, lung, colorectal, prostate, ovarian, liver, pancreatic, and stomach), using data from the All of Us Research Program, which integrates EHR, genomic, and survey data from over 865,000 participants. Even with a baseline modeling approach, EHR-based models achieved a 3- to 6-fold higher enrichment of true cancer cases among individuals identified as high risk compared with traditional risk factors alone, whether used as a standalone or complementary tool. The EHR foundation model, a state-of-the-art approach trained on comprehensive patient trajectories, further improved predictive performance across 26 cancer types, demonstrating the clinical potential of EHR-based predictive modeling to support more precise and scalable early detection strategies.

## Introduction

Early detection of cancer, before it progresses to advanced stages, can substantially improve survival (e.g., 44% vs. 3% five-year survival for early- vs. late-stage pancreatic cancer[1]) and reduce cancer-related mortality.[2] However, effective screening guidelines currently exist for only a few cancer types[3], such as colorectal cancer (starting at age 45)[4] and lung cancer (based on smoking history)[5]. Many cancers with high case-fatality rates, such as pancreatic, liver, ovarian, and stomach cancers, lack evidence-based screening strategies. These cancers are often diagnosed at advanced stages, due to their insidious onset, low symptom specificity, and absence of effective early detection tools. Notably, they are among the leading causes of cancer-related mortality[6,7], underscoring the urgent need for scalable risk-tailored approaches to identify high-risk individuals who may benefit from early diagnostic interventions.

Electronic health records (EHRs) offer a promising, non-invasive, and cost-effective data source for identifying high-risk individuals across diverse cancer types.[8] EHRs capture longitudinal patient trajectories that may reveal prediagnostic symptom clusters or healthcare utilization patterns, spanning diagnoses, medications, procedures, and more, enabling personalized risk profiling.[9] Recent advances in artificial intelligence (AI), particularly the emergence of large language models (LLMs) capable of synthesizing large-scale clinical data, have further amplified the potential of EHR-based predictive models.[10-12]

The recent development of EHR foundation models, trained on longitudinal patient trajectories analogous to how LLMs such as ChatGPT are trained on vast text corpora, has demonstrated strong transferability and state-of-the-art performance across diverse clinical tasks.[13-17] These advances suggest the feasibility of a unified, scalable, and generalizable EHR-based early detection framework applicable to multiple cancer types.

However, many existing EHR-based cancer risk models have focused primarily on methodological innovations (e.g., gradient-boosted trees, deep neural networks, and transformer architectures) and their predictive performance for single cancer types, with limited evaluation of clinical utility. [18-20] In particular, there has been less attention on how these models perform in stratifying high-risk populations for targeted screening or how they compare with traditional risk factors such as family history, genetic carrier status, or relevant comorbidities.

To address this gap, we evaluated the performance of EHR-based predictive models in identifying high-risk individuals compared with traditional risk factors, including those currently used in screening guidelines or considered in clinical trials.[21,22] We focused on risk enrichment, how effectively each approach identifies high-risk cohorts with higher true cancer prevalence relative to the general population (also referred to as *lift* values) across eight major cancer types. Using data from the All of Us Research Program, a national cohort of more than 865,000 participants integrating longitudinal EHR, genomic, and survey data[23], we directly compared EHR-based predictions with well-established risk factors, including age, family history, and pathogenic genetic variants[24,25] (e.g., *BRCA*, *Lynch syndrome*).

Our results show that even baseline models such as XGBoost can identify high-risk cohorts more effectively than traditional risk factors, either as standalone tools or in combination with them. Incorporating EHR foundation models further enhanced predictive performance across multiple cancer types. This large-scale, multi-cancer evaluation provides an important framework for advancing early cancer detection through the use of EHR data.

## Results

### Study workflow and cancer cohort identification

The overall study workflow, including cohort identification, predictive modeling, and clinical utility evaluation against established risk factors, is summarized in Figure 1. We used structured EHR data harmonized under the Observational Medical Outcomes Partnership (OMOP) common data model.[26,27]

To identify cancer cohorts, we developed a prompt-based classification approach using OpenAI's GPT-4o large language model via the API[28] to categorize approximately 54,000 malignancy-related condition names under the highest-level OMOP concept, "Malignant neoplastic disease". Through iterative refinement with clinical review, we defined 52 cancer categories, prioritizing tissue of origin over anatomical location (Table S1).

Applying the final prompt to 796 unique cancer-related terms (a subset of 54,000) identified in the Columbia University Irving Medical Center (CUIMC) dataset achieved 94.4% mapping accuracy, while application to 999 terms in the All of Us (AoU) dataset achieved 98.4% accuracy. Misclassifications identified during this process were corrected.

In the AoU database, we identified about 63,000 individuals with confirmed cancer diagnoses and 570,000 controls without malignancy related conditions. In the CUIMC database, we identified about 500,000 cancer cases and 4.3 million controls. Cancer cases were classified into 50 distinct cancer types in AoU and 49 types in CUIMC. After the preprocessing steps for cancer cohort identification described in the Methods, the final analytic cohorts included approximately 38,000 cancer cases and 210,000 controls from AoU, and 120,000 cancer cases and 1.1 million controls from CUIMC (Table S2). Because the CUIMC data do not include genomic or survey information, we conducted the clinical utility evaluation of EHR-based models using AoU data only, while assessing EHR foundation model performance with both CUIMC and AoU data.

### Performance of EHR-based baseline models across cancer types

We developed cancer-specific predictive models using XGBoost[29], leveraging medical conditions documented at least 12 months prior to diagnosis. Each model incorporated approximately 27,000 features derived from EHR. Predictive performance, measured by the area under the receiver operating characteristic curve (AUROC), is shown in Figure 1 for 26 cancer types with at least 100 diagnosed individuals. XGBoost models generally performed better than logistic regression models. To ensure comparability, we evaluated each model against the same control group described in the cohort definition. We used five-fold stratified cross-validation, with test sets representing hypothetical general populations for evaluating model performance against established risk factors. Performance varied across cancer types, with the highest AUROC for prostate cancer (0.90; 95% CI: 0.89–0.90) and the lowest for stomach cancer (0.59; 95% CI: 0.55–0.63).

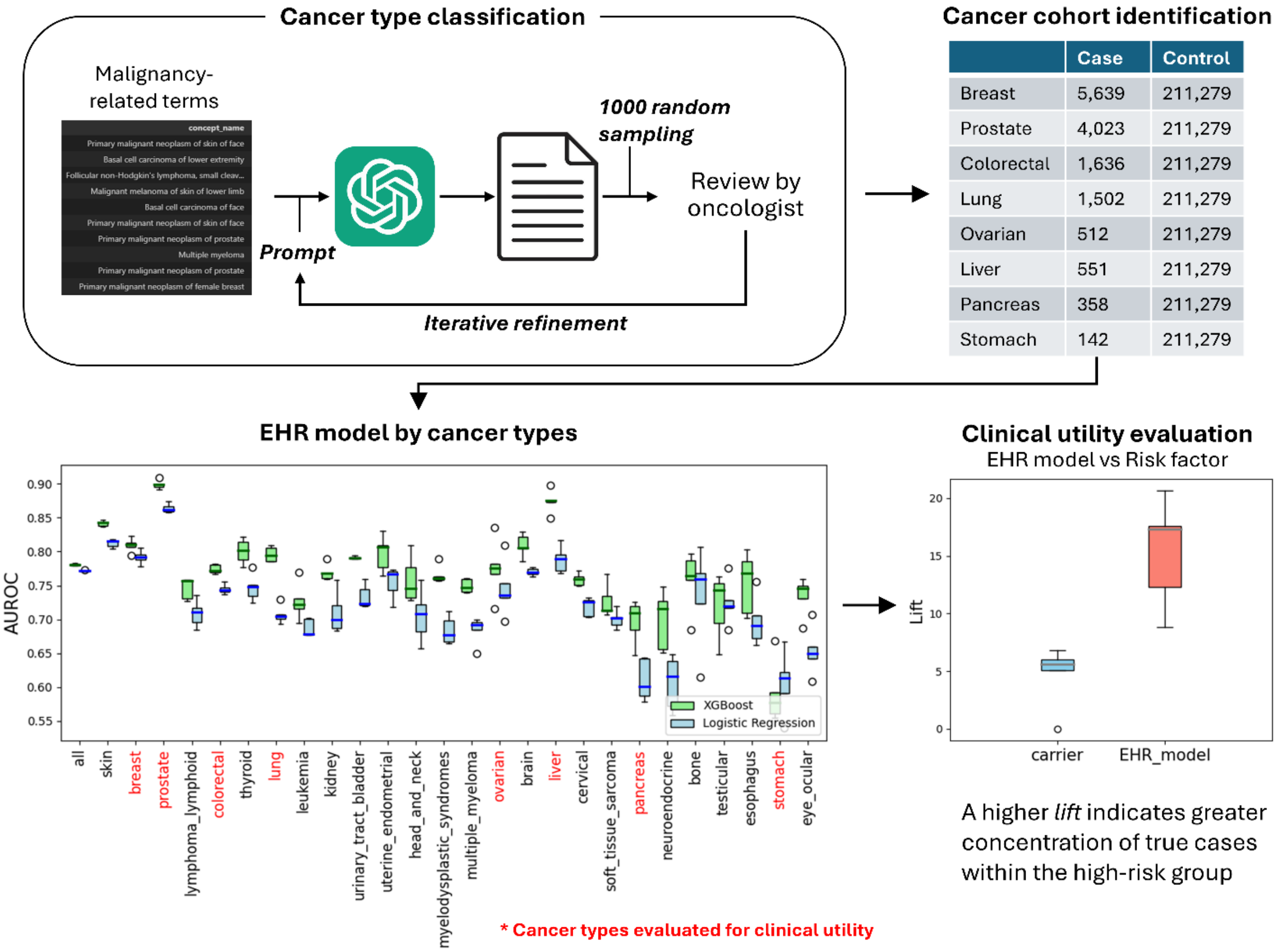

**Figure 1. Study workflow for evaluating clinical utility of EHR model compared to established risk factors.** Among 50 cancer types identified in the All of Us data, AUROC distributions are shown for cancer types with at least 200 cases. Cancer types in red (e.g., pancreas, colorectal, breast) were selected for clinical utility comparison. For example, comparison of EHR-based model and traditional risk factor (e.g., carrier status) for pancreatic cancer, measured using *lift* at equal population coverage (1%).

## Clinical utility of EHR-based models in comparison with traditional risk factors

To quantify how well each criterion (e.g., EHR-based model, age, *BRCA*) enriches the high-risk cohort relative to the general population, we used the metric *lift*,[30,31] defined as the ratio of cancer prevalence in the high-risk cohort to that in the overall population. A higher *lift* indicates a greater concentration of true cases within the high-risk group, making the approach more effective for diagnostic evaluation (e.g., imaging or biopsy). The proportion of the identified high-risk group within the general population is referred to as coverage (%). The EHR model, which assigns individual risk scores, defines the high-risk group using a predefined threshold corresponding to a specific population coverage. To compare EHR-based models with traditional risk factors (e.g., age, gene mutations), we measured *lift* at equal population coverage determined by each traditional risk factor (Table 1). For example, in pancreatic cancer, which currently lacks standard screening guidelines, the ongoing Cancer of the Pancreas Screening Study (CAPS) trial[20] defines high-risk individuals based on the presence of inherited mutations (e.g., *ATM*, *BRCA1/2*, *PALB2*, *STK11*, *PRSS1/2*, *CTRC*, and *Lynch syndrome* genes) or a family history of pancreatic cancer (two or more close relatives on the same side of the family). In the AoU cohort, individuals carrying one of these gene mutations represented approximately 1% of the population, and had a *lift* of 4.71, meaning a 4.71-fold higher concentration of true cases than in general population. At the same coverage, the EHR-based model achieved a *lift* of 15.3 (Table 1), demonstrating substantially greater enrichment of high-risk individuals than using genetic carrier status alone. By contrast, a family history of any cancer yielded a *lift* of 0.78 (below 1), indicating poorer performance than screening the general population. When limited to family history of pancreatic cancer, the *lift* increased to 4.12. We also evaluated individuals with new-onset diabetes (NOD), with and without an age constraint of 60 years. Across all these scenarios, the EHR-based model demonstrated superior performance in identifying enriched high-risk cohorts (Figure 2A).

In addition to evaluating the clinical utility of the EHR model as a standalone approach, we assessed its effectiveness when combined with current clinical guidelines or known risk factors (Figure 2B). Because combining two criteria expands the high-risk group, it tends to reduce the overall *lift* compared with using the EHR model alone, as the broader group likely includes more false positives. Nonetheless, our results show that integrating the EHR model with established risk factors still significantly improves *lift* compared with using the risk factor alone, while also broadening the high-risk pool. For example, combining carrier status with the EHR model increased *lift* to 6.74 while expanding coverage to 3%, compared with a *lift* of 4.71 at 1% using carrier status alone (Figure 2B).

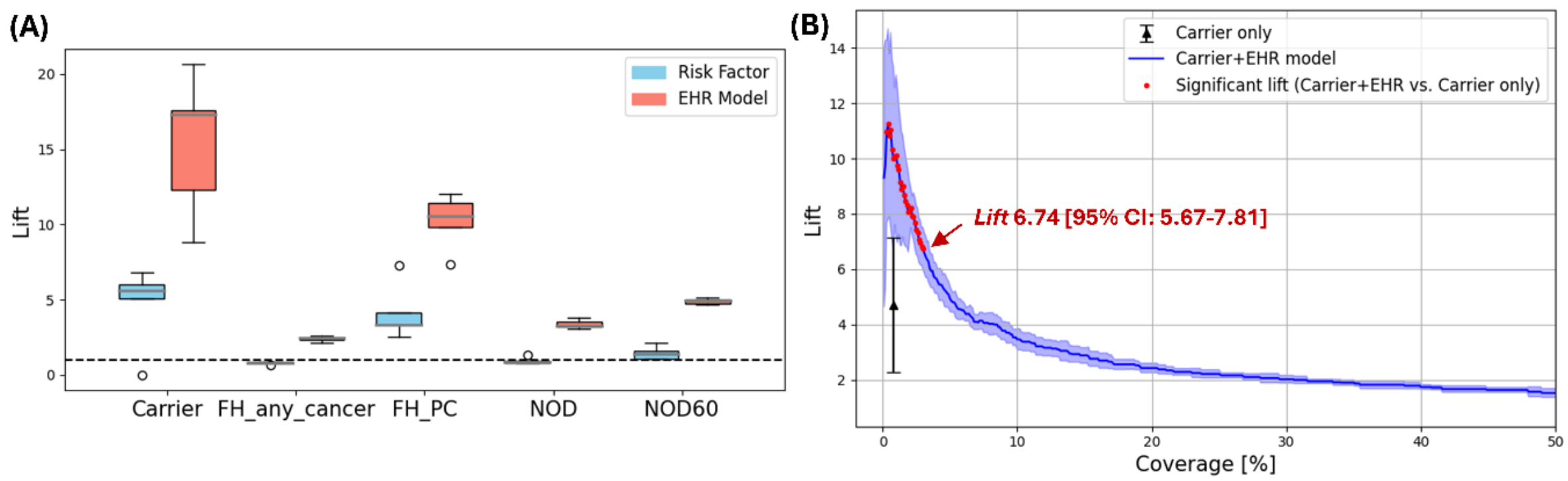

**Figure 2. Clinical utility evaluation of the EHR model for pancreatic cancer.** (A) Clinical utility of the EHR model as an independent tool: *lift* achieved by the EHR model compared to traditional risk factors at matched coverage levels. (B) Clinical utility of the EHR model as a complementary tool: *lift* when combining the EHR model with a risk factor to identify high-risk individuals, evaluated across a range of targeted coverage thresholds. The added value of the EHR model varies depending on the coverage used to define high-risk groups. Thresholds yielding statistically significant lift (EHR + risk factor vs. risk factor alone) are highlighted in red dots.

Beyond pancreatic cancer, we evaluated the model's clinical utility for breast, prostate, colorectal, lung, ovarian, liver, and stomach cancers (Table 1). In most cases, the EHR model significantly improved *lift*, both as a standalone tool and in combination with established risk factors. There were two cases where the EHR model was not more effective. The first was colorectal cancer, where performance was similar to the current screening guideline of age 45. *Lift* from the EHR model was comparable to that achieved using the age-based criterion. However, unlike age-based screening, the EHR model is not constrained by age thresholds, offering the potential to identify early-onset colorectal cancers, which have shown a concerning rise in incidence and mortality in recent years.[32] In this case, combining the EHR model with the age criterion as a complementary approach may provide optimal screening utility. The second exception was stomach cancer, where the lift achieved by the EHR model (4.43, 95% CI: 0.81–8.05) was lower than that based on family history of stomach cancer (6.64). However, this difference should be interpreted with caution due to the wide confidence intervals, likely reflecting the limited sample size for this subgroup.

**Table 1. Clinical utility evaluation across cancer types.** *Lift* values from the EHR-based model were evaluated at population coverages matched to those of the traditional risk factors (RF), demonstrating its value as an independent tool for identifying high-risk individuals. *Lift* values from the combined RF + EHR model illustrate its value as a complementary tool when used alongside traditional risk factors. * indicates statistical significance at $p < 0.05$.

| | Risk Factor (RF) | Coverage | Lift by RF | Lift by EHR model | Max lift by RF+EHR model |
|---|---|---|---|---|---|
| **Breast** | BRCA | 0.4% | 5.90 [4.84-6.97] | 21.6 [20.1-23.1]* | 13.6 [12.9-14.3]* |
| | Age 40-74 | 64% | 1.32 [1.31-1.34] | 1.44 [1.43-1.45]* | 1.37 [1.35-1.39]* |
| | FH_cancer | 20% | 1.27 [1.18-1.36] | 3.29 [3.13-3.45]* | 2.25 [2.14-2.36]* |
| | FH_breast | 8.0% | 2.61 [2.44-2.78] | 5.51 [5.06-5.96]* | 4.11 [3.90-4.32]* |
| **Prostate** | Age 55-69 | 30% | 1.84 [1.77-1.92] | 2.95 [2.90-3.00]* | 2.37 [2.30-2.44]* |
| | FH_cancer | 20% | 1.26 [1.20-1.32] | 4.12 [4.04-4.20]* | 2.72 [2.62-2.81]* |
| | FH_prostate | 4% | 4.52 [4.24-4.80] | 11.2 [10.8-11.7]* | 7.77 [7.50-8.04]* |
| **Colorectal** | Age 45 | 64% | 1.39 [1.33-1.44] | 1.38 [1.34-1.41] | 1.41 [1.36-1.46] |
| | FH_cancer | 20% | 0.98 [0.89-1.07] | 3.09 [3.02-3.16]* | 2.16 [2.08-2.24]* |
| | FH_colorectal | 5% | 2.70 [2.46-2.93] | 7.72 [7.45-7.99]* | 5.74 [5.35-6.14]* |
| **Lung** | Smoking history | 39% | 1.78 [1.71-1.85] | 2.03 [1.97-2.09]* | 1.88 [1.82-1.94]* |
| | Smoking history & Age 50-80 | 23% | 2.59 [2.44-2.75] | 2.89 [2.76-3.03]* | 2.73 [2.67-2.79]* |
| | FH_cancer | 20% | 1.04 [0.95-1.13] | 3.27 [3.18-3.36]* | 2.18 [2.07-2.28]* |
| | FH_lung | 5% | 2.80 [2.28-3.33] | 7.22 [6.47-7.96]* | 4.92 [4.59-5.25]* |
| **Ovarian** | BRCA | 0.4% | 8.82 [3.18-14.5] | 25.1 [15.1-35.1]* | 18.9 [13.7-24.0]* |
| | FH_cancer | 20% | 1.25 [1.07-1.44] | 3.19 [2.64-3.74]* | 2.33 [2.19-2.46]* |
| | FH_ovarian | 2% | 7.39 [6.67-8.12] | 11.0 [9.64-12.3]* | 10.6 [9.22-11.9]* |
| **Liver** | Hepatitis B/C or Cirrhosis | 3% | 15.9 [14.4-17.5] | 22.2 [20.7-23.8]* | 18.2 [16.8-19.7]* |
| | FH_cancer | 20% | 0.70 [0.56-0.84] | 4.48 [2.97-5.99]* | 2.96 [2.76-3.16]* |
| **Pancreas** | carrier | 1% | 4.71 [1.35-8.07] | 15.3 [9.50-21.2]* | 11.2 [6.47-16.0]* |
| | FH_cancer | 20% | 0.78 [0.67-0.89] | 2.39 [2.17-2.60]* | 1.65 [1.44-1.86]* |
| | FH_pancreas | 2% | 4.12 [1.81-6.43] | 10.2 [7.97-12.5]* | 7.73 [5.13-10.3]* |
| | NOD | 13% | 0.95 [0.67-1.23] | 3.39 [3.03-3.74]* | 2.34 [2.25-2.43]* |
| | NOD60 | 6% | 1.44 [0.90-1.99] | 4.88 [4.65-5.12]* | 3.71 [3.61-3.82]* |
| **Stomach** | Helicobacter pylori | 2% | 1.43 [0.00-4.28] | 4.04 [1.91-6.17]* | 3.33 [2.87-3.79] |
| | FH_cancer | 20% | 0.83 [0.33-1.33] | 1.67 [1.49-1.85]* | 1.45 [1.21-1.69]* |
| | FH_stomach | 1% | 6.64 [3.49-9.79] | 4.43 [0.81-8.05] | 6.99 [4.61-9.38] |

## EHR foundation model performance

We evaluated two EHR foundation models, CEHR-GPT[33] and MOTOR[34], which were developed using distinct learning objectives[35], and compared their performance with baseline models (e.g., XGBoost and logistic regression models) (Figure S2). MOTOR, previously shown to achieve state-of-the-art performance among representation learning approaches for various downstream predictions (e.g., via linear probing)[36], also demonstrated strong performance in our cancer risk prediction tasks.

For clinical utility evaluation (Table 1), we used XGBoost model's 1-year-prior risk predictions based on condition data only. Adding data from medication, procedure, and demographic domains significantly improved performance across all eight major cancers (Table 2). With the foundation models, performance improved even further. Notably, the foundation model's 3-year-prior risk predictions outperformed the 1-year predictions generated by the baseline XGBoost model used in the clinical utility evaluation. Even after expanding the baseline model to include additional domains, CEHR-GPT's 3-year performance remained comparable to the baseline 1-year performance, whereas MOTOR continued to outperform in many cancers, including breast, colorectal, lung, pancreas, and liver. EHR foundation models demonstrated stronger performance on CUIMC data than on AoU data across 26 cancer types (Figure S2). The CUIMC dataset contained roughly three times more samples, whereas the AoU dataset was approximately three times richer in sequence length, while the number of unique concept IDs was similar between the two datasets (Table S2).

**Table 2. Performance of EHR foundation models compared with baseline models for 1- and 3-year risk prediction.** Baseline models refer to XGBoost. * indicates statistical significance at $p < 0.05$ when comparing baseline 1-year–prior predictions (including condition, medication, procedure, demographics features) with EHR foundation models' 3-year–prior predictions.

| | Baseline | | Baseline | | CEHR-GPT | MOTOR |
|---|---|---|---|---|---|---|
| Domains | condition | condition | condition, medication, procedure, demographics | condition, medication, procedure, demographics | condition, medication, procedure, demographics | condition, medication, procedure, demographics, observation, lab |

| Prediction time prior to diagnosis | 1year | 3year | 1year | 3year | 3year | **3year** |
|---|---|---|---|---|---|---|
| Breast | 0.810 [0.797, 0.822] | 0.794 [0.791, 0.798] | 0.872 [0.862, 0.882] | 0.866 [0.856, 0.877] | 0.883 [0.872, 0.893] | **0.901 [0.891, 0.910]*** |
| Prostate | 0.899 [0.891, 0.907] | 0.884 [0.878, 0.891] | 0.945 [0.939, 0.950] | 0.942 [0.936, 0.947] | 0.946 [0.939, 0.952] | **0.948 [0.942, 0.954]** |
| Colorectal | 0.774 [ 0.765, 0.783] | 0.763 [0.746, 0.779] | 0.818 [0.790, 0.844] | 0.806 [0.779, 0.833] | 0.827 [0.798, 0.852] | **0.878 [0.854, 0.901]*** |
| Lung | 0.796 [0.782, 0.810] | 0.770 [0.757, 0.784] | 0.820 [0.793, 0.844] | 0.808 [0.782, 0.835] | 0.840 [0.814, 0.862] | **0.869 [0.847, 0.890]*** |
| Ovarian | 0.775 [0.723, 0.828] | 0.751 [0.727, 0.775] | 0.864 [0.827, 0.900] | 0.856 [0.820, 0.893] | 0.846 [0.803, 0.887] | **0.860 [0.810, 0.898]** |
| Pancreas | 0.697 [0.658, 0.737] | 0.655 [0.605, 0.706] | 0.749 [0.689, 0.805] | 0.732 [0.675, 0.790] | 0.796 [0.748, 0.839] | **0.845 [0.792, 0.884]*** |
| Liver | 0.847 [0.853, 0.895] | 0.846 [0.811, 0.882] | 0.891 [0.850, 0.927] | 0.878 [0.829, 0.921] | 0.914 [0.885, 0.938] | **0.933 [0.904, 0.958]*** |
| Stomach | 0.590 [0.534, 0.647] | 0.637 [0.563, 0.711] | 0.685 [0.585, 0.778] | 0.659 [0.536, 0.759] | 0.774 [0.670, 0.857] | **0.798 [0.689, 0.902]** |

## Discussion

Our findings address a critical gap in early cancer detection, where only a few cancer types currently have established screening guidelines, and most rely on narrow eligibility criteria such as age or smoking history. For cancers lacking effective screening strategies such as pancreatic, liver, ovarian, and stomach cancers, early diagnosis remains challenging despite clear survival benefits when detected at an early stage. By evaluating EHR-based predictive models across multiple cancer types, we demonstrate that routinely collected clinical data can help identify high-risk individuals beyond traditional risk factors and current guideline criteria.

The models effectively stratified high-risk populations, either as a standalone tool or in combination with existing risk criteria. Even the baseline model built using only the condition domain improved the identification of high-risk groups, achieving greater enrichment of true positives (referred to as *lift*) compared with traditional risk factors, and potentially enabling more effective diagnostic evaluation when triaged to imaging or biopsy. The EHR foundation model demonstrated additional gains in performance (Table 2), further highlighting the potential of EHR-based approaches to transform early cancer detection and advance population-level cancer prevention in the era of AI-driven precision medicine.

Our evaluation of *lift* relative to traditional risk factors provides a framework for determining optimal thresholds to identify high-risk individuals using EHR models for each cancer type. We plan to first define candidate threshold ranges and then identify the optimal threshold within those ranges through counterfactual simulation. For example, in pancreatic cancer, the highest *lift* value from a known risk factor (family history of pancreatic cancer) was 4.10 at 2% coverage, whereas the baseline EHR model achieved a *lift* of 8.54 at the same coverage, and we anticipate that the foundation model will perform even better. Because *lift* typically decreases as coverage increases, we will identify the maximum coverage at which *lift* remains ≥4.10, thereby defining a candidate threshold range (e.g., 2% to X%) for further evaluation. Within this range, we will conduct counterfactual retrospective analyses to quantify the number of patients flagged, the number and timing of true cancers detected, false positive rates, the estimated number of screening tests and downstream workups, clinical and resource implications, and potential stage shifts associated with each threshold. By systematically assessing these metrics, we aim to identify the target coverage (flagging threshold) for each cancer type.

To assess key features contributing to model predictions, we used SHAP[37], a widely used feature-attribution method commonly applied in machine learning and deep learning studies. Across cancer types, the top-ranked features varied but often included known risk factors or clinically related conditions (Figure S1). For example, cirrhosis and hepatitis were among the most influential features for liver cancer, while pancreatic cysts and other pancreatic disorders ranked highly for pancreatic cancer. However, given the superior performance of the EHR model over these traditional risk factors in our *lift* evaluation, its improved risk stratification likely reflects the ability to capture not only known associations, but also additional, non-obvious features identified in the SHAP analysis. This demonstrates how EHR-based models can leverage a broad spectrum of structured clinical data to identify meaningful predictors of cancer risk, many of which may be overlooked by rule-based or linear approaches.

In the breast cancer model, top predictive features included "carcinoma in situ of the breast" and "acquired absence of breast," even in predictions made three years prior to diagnosis. Although "carcinoma in situ" is noninvasive and "acquired absence of breast" does not directly indicate active cancer, these findings highlight the importance of validating cancer diagnoses and their timing derived from structured EHR data. Because unstructured data (e.g., clinical notes), often considered the gold standard for confirming cancer cases, are frequently unavailable, as in the All of Us dataset, or difficult to share across institutions, developing optimized case definitions based on structured data is essential for building effective and generalizable EHR-based models.

These findings reflect inherent limitations of EHR-based research, including data incompleteness, misclassification, and potential false positives arising from documentation or coding errors.[38,39] While manual chart review remains important for ensuring phenotypic accuracy, HIPAA compliant, LLM-based tools, such as the GPT prompting approach used in this study to classify 54,000 concept names into specific cancer types, may enable scalable, semi-automated validation pipelines that support data curation and harmonization at scale.

Although SHAP and other attribution methods[40] are widely used, interpreting their outputs remains challenging because feature importance can be affected by noise or bias in the underlying data and presented in ways that fail to earn clinician trust. For foundation models, interpretability becomes even more complex. Applying traditional feature attribution methods to these models is nearly infeasible given their scale and architecture, and robust interpretability frameworks are not yet well established. A promising direction involves leveraging recent advances in generative AI's reasoning capabilities to develop clinician-facing interfaces that allow users to interact with models and explore how predictions are generated. Such tools could provide contextual, human-understandable explanations and improve transparency in model decision-making. As AI technologies continue to advance and become increasingly integrated into clinical care, developing explainable AI methods that enhance human understanding will be essential for the responsible translation of predictive models into practice.

While our study demonstrates the potential of EHR-based models to improve early cancer detection, any predictive tool used to guide screening decisions must be evaluated in the context of overdiagnosis and overtreatment.[2] Identifying individuals at elevated risk does not necessarily translate to clinical benefit, particularly for cancers with indolent or slow-progressing courses. Future work should incorporate downstream clinical outcomes to assess whether EHR-based risk stratification leads to net benefit in terms of survival, quality of life, and healthcare resource utilization.

Another key consideration is potential bias in model development and deployment. Because the AoU cohort is based on voluntary participation, selection bias may influence the representativeness of the data despite the use of internal controls. Moreover, if EHR-based models were deployed in practice, differences in diagnostic intensity or healthcare access across demographic and socioeconomic groups could affect model calibration and performance, potentially leading to new inequities if unaddressed. Continuous bias monitoring, recalibration, and evaluation across subpopulations will therefore be essential to ensure that EHR-based predictive models promote equity and generalizability in real-world clinical use. As part of our ongoing work, we plan to perform multi-site fine-tuning of the foundation model to evaluate and mitigate site-specific biases and enhance model generalizability across diverse healthcare systems.

Additionally, traditional risk factors such as family history are self-reported in the AoU dataset and may be subject to underreporting, as noted in prior studies.[41] This limitation may lead to underestimation of the predictive value of conventional risk factors relative to EHR-based models.

In conclusion, EHR-based predictive models offer a non-invasive, scalable approach to identifying individuals at elevated cancer risk, with the potential for meaningful clinical impact in early detection. However, this promise must be balanced with comprehensive evaluation of downstream effects to ensure a net benefit in real-world applications. With continued improvements in data quality, model interpretability, and integration of diverse data types, these tools can complement existing screening guidelines and inform personalized surveillance strategies. Future work should explore how high-risk individuals identified by such models can be enrolled in risk-adapted screening protocols, including decisions around eligibility, screening frequency, and modality. These efforts represent a critical step toward realizing precision prevention at a population scale.

## Methods

### Database

We used databases from two sources (Table 3): Columbia University Irving Medical Center (CUIMC) and the All of Us (AoU) Research Program (Controlled Tier Dataset, version 8), both standardized to the OMOP common data model. CUIMC is a large academic medical center serving a diverse urban population, with longitudinal inpatient and outpatient EHR data including demographics, conditions, medications, laboratory measurements, procedures, and visit information collected as part of routine clinical care.

The AoU Research Program[23] provides a nationally representative dataset designed to reflect the diversity of the U.S. population. Any individual aged 18 or older is eligible to participate. As of October 2025, more than 865,000 participants have enrolled in the program, with over 595,000 completing full consent, including EHR linkage, surveys, physical measurements, and biospecimen donation.

### Cancer type classification

We extracted all descendant concept names (~54,000) under the OMOP concept ID 443392, which represents malignant neoplastic disease. We then developed a GPT-based classification approach to map these medical concept names to predefined cancer types (e.g., colorectal, liver, or skin cancer).

Through an iterative process with clinician review (Figure 1), both the predefined cancer site list and the classification prompt were refined. The ~54,000 cancer-related concept names under concept ID 443392 included terms with a clearly specified tissue of origin as well as those with unclear or unspecified origin. Our prompt prioritized tissue of origin over anatomical location whenever possible; if the tissue of origin was not explicitly stated, the anatomical location mentioned in the concept name was used instead.

Consequently, the resulting 52-site schema included a mix of categories defined by tissue of origin (e.g., ovarian or brain) and those defined by anatomical or histologic context (e.g., choriocarcinoma) (Table S1). Classification was performed using the OpenAI API with the GPT-4o model, and the final prompt is publicly available at https://github.com/jp4147/aou_EHRmodel.

### Cancer cohort identification

We identified individuals with any concept IDs corresponding to malignant neoplastic disease (i.e., descendants of OMOP concept ID 443392) recorded in their medical history. The earliest occurrence of any such concept ID was defined as the individual's first cancer diagnosis date. Individuals without any cancer-related concept IDs were considered potential controls.

For cases, the 1-year prediction index date was defined as exactly 12 months prior to the first recorded cancer diagnosis and was treated as a prospective prediction time point. To prevent information leakage from diagnostic workup or early cancer-related symptoms, all model features were restricted to events occurring strictly before the index date. No clinical data recorded on or after the index date were used for model input.

For controls, the index date was set to 24 months prior to the last recorded medical condition to reduce the likelihood of including individuals with undiagnosed cancer, consistent with prior EHR-based cancer risk prediction studies[42,43] For the 3-year prediction setting, the index date was shifted two years earlier relative to the 1-year index date for both cases and controls.

To ensure adequate longitudinal medical history, we included only individuals (cases and controls) with at least five documented medical conditions prior to their respective index dates, operationalized as at least five non-duplicate condition occurrences recorded before the index date (after removing duplicate entries for the same condition recorded on the same date). We selected a threshold of five as a pragmatic minimum to exclude individuals with extremely sparse longitudinal documentation while retaining broad cohort coverage.

**Risk factor identification**

We evaluated a set of established risk factors to compare against the EHR model, including: age, family history of cancer, smoking history, genetic carrier status, chronic hepatitis B/C, cirrhosis, new-onset diabetes, and *Helicobacter pylori* infection.

Age was calculated relative to each individual's index date using their recorded date of birth. For conditions such as chronic hepatitis B/C, cirrhosis, new-onset diabetes, and *H. pylori* infection, we queried AoU database using curated sets of high-level OMOP concept IDs associated with each condition. We retrieved all descendant concepts via the cb_criteria and cb_search_all_events tables and flagged individuals with any matching concept recorded in their medical history. Individuals were classified as having each risk factor if they had a first recorded diagnosis prior to their index date.

For new-onset diabetes, individuals were classified as positive if they had a first diagnosis of type 2 diabetes before the index date and no record of prior diabetes-related medications. Family history of cancer and smoking history were derived from participant survey data.

Genetic carriers were identified using the ClinVar database. We restricted the analysis to variants classified as *pathogenic* or *likely pathogenic*, submitted by multiple submitters with no conflicts, or reviewed by an expert panel. A total of 4,738 individuals were identified as carriers of pathogenic variants in genes associated with cancer predisposition syndromes, including *ATM*,

*BRCA1*, *BRCA2*, *PALB2*, *STK11* (Peutz-Jeghers), *CDKN2A* (FAMMM), and Lynch syndrome genes (*EPCAM*, *MLH1*, *MSH2*, *MSH6*, *PMS2*), as well as hereditary pancreatitis genes (*PRSS1/2*, *CTRC*).

**Model training and evaluation setup**

To ensure a consistent modeling framework across all cancer types, we used a shared control group of 211,279 individuals. For the baseline EHR models used in the clinical utility evaluation, we intentionally included only condition-based features and excluded demographic variables to isolate and evaluate the predictive signal contained within clinical condition data alone. This design allowed us to assess the contribution of EHR-derived features independent of demographic information.

For the EHR foundation models, which were trained on the entire EHR and implicitly capture demographic context, we did not construct separate sex-specific control groups (e.g., for prostate or ovarian cancer). Although sex-specific controls could further refine performance estimates for sex-limited cancers, we selected this unified approach to maintain comparability and to evaluate the generalizability of EHR-based modeling across cancer types.

*Baseline models for clinical utility evaluation*

We included only condition-based features to focus on evaluating the predictive signal contained within existing clinical condition data alone. Unknown concepts (i.e., concept ID = 0) and duplicate entries (e.g., the same condition recorded on the same date) were excluded.

We conducted five-fold stratified cross-validation separately for each cancer type. While the number of cases varied by cancer type, we used a consistent control group across all models. In each fold, the data were split into a training set (80%) and a test set (20%), maintaining the case-control ratio.

For each individual, medical condition data were converted into a binary sparse matrix, where each feature represented the presence or absence of a specific condition prior to the index date. Using these features, we trained gradient-boosted decision tree models with XGBoost for each fold.

While the training sets were used to develop the EHR models, the test sets served as a proxy for a hypothetical general population to evaluate the clinical utility of each model. We compared the effectiveness of the EHR model and known risk factors in identifying high-risk individuals.

To quantify clinical utility, we used *lift*, a metric that measures the enrichment of true cases within the high-risk group relative to the general population. A higher lift indicates that the high-risk group contains a greater concentration of true cancer cases.

*EHR foundation models*

Unknown concepts (i.e., concept ID = 0) were excluded from all domains except the visit type when constructing patient sequences proposed in our previous study.[33] Additionally, patients with fewer than 20 tokens were excluded to ensure sufficient longitudinal information for model training. The final CUIMC dataset included approximately 2.6 million patients for training and 1 million patients for evaluation[33], while the AoU dataset included approximately 255,000 patients for training and 64,000 patients for evaluation.

We evaluated two existing EHR foundation model frameworks, CEHR-GPT[33] and MOTOR[34], for their predictive performance in cancer risk prediction across 26 cancer types using a linear probing approach. CEHR-GPT was developed at CUIMC and MOTOR at Stanford University. Although MOTOR is compatible with OMOP data, the pretrained MOTOR model was trained on a different data source, so we could not directly fine-tune it on CUIMC data. Therefore, we retrained MOTOR from scratch using CUIMC and AoU data.

We pretrained the two models on preprocessed data from approximately 2.6 million CUIMC patients and 255,000 All of Us (AoU) participants, respectively. CEHR-GPT incorporates data from the demographics, condition, medication, and procedure domains, while MOTOR additionally includes laboratory and observation data. For baseline comparison models (e.g., XGBoost and logistic regression), we used the same domains as CEHR-GPT (demographics, condition, medication, and procedure).

MOTOR is a non-generative foundation model trained with time-to-event prediction objectives, achieving state-of-the-art performance across diverse downstream tasks via linear probing. In contrast, CEHR-GPT is a generative model trained on next-token prediction, enabling both linear probing and zero-shot inference.[33-35]

The initial CEHR-GPT architecture was designed around the condition, procedure, and drug domains to align with Observational Health Data Sciences and Informatics (OHDSI) studies[44], which primarily use claims data that lack laboratory information. Although we have also trained CEHR-GPT variants incorporating lab measurement and observation domains, their contributions to prediction performance were marginal. We hypothesize that information from these domains is largely correlated with condition/procedure/drug data, and that naively appending all data points can degrade model performance. For instance, MOTOR applies entropy filtering to exclude non-informative codes, reducing its vocabulary from ~50,000 to ~9,000 concepts.

We are continuously refining CEHR-GPT and plan to release updated versions in future studies. Our long-term goal is to achieve representation performance comparable to MOTOR while retaining CEHR-GPT's generative capabilities. Detailed implementation and comparison results are available in preprints. The full training code for CEHR-GPT and MOTOR on OMOP data, including an AoU tutorial, is publicly available at https://github.com/knatarajan-lab/cehrgpt.

### Feature importance extraction

To identify features contributing to model predictions, we computed SHAP (SHapley Additive exPlanations) values for each individual in the test set across all five cross-validation folds. We hypothesized that features whose SHAP values show inconsistent directionality across patients may reflect unreliable contributions to risk prediction. Accordingly, for each fold, we aggregated SHAP values by summing the signed contributions of each feature across individuals. Features were then ranked within each fold based on their aggregate contributions, and final rankings were obtained by averaging ranks across folds to identify consistently important predictors.

### Statistical analysis

We reported 95% confidence intervals (CIs) to represent variability in performance metrics across the five cross-validation folds. For EHR foundation model performance, 95% CIs were estimated using bootstrap resampling.

For the clinical utility evaluation, we compared lift values between the EHR-based model and traditional risk factor–based stratification using the one-sided Mann-Whitney U test (alternative = 'less'), assessing whether the EHR model achieved significantly higher lift.

For comparisons between baseline and foundation models (e.g., XGBoost vs. CEHR-GPT or MOTOR), we performed one-sided bootstrap tests (1,000 resamples) comparing distributions of performance metrics across cancer types. Empirical P-values were computed as the proportion of bootstrap samples in which the performance difference (foundation - baseline) was $\leq 0$. P-values $< 0.05$ were considered statistically significant.

### Author contributions statement

J.P. led the study design, performed all analyses and experiments, and drafted the manuscript. C.P. contributed to the design and interpretation of the foundation model experiments. T.L., J.Y., and A.W. assisted with prompt development for the cancer-type classification pipelines and provided clinical expertise. N.T. and J.B. contributed to the conception and design of the work and to the interpretation of results. All authors reviewed the manuscript, provided comments, and approved the final version.

**Funding statement**

This work was supported by National Cancer Institute of the National Institutes of Health under K25CA267052 (J.P.)

## References

1. Bestari, M.B., Joewono, I.R. & Syam, A.F. A Quest for Survival: A Review of the Early Biomarkers of Pancreatic Cancer and the Most Effective Approaches at Present. *Biomolecules* **14**(2024).
2. Crosby, D., *et al.* Early detection of cancer. *Science* **375**, eaay9040 (2022).
3. Smith, R.A., *et al.* Cancer screening in the United States, 2016: A review of current American Cancer Society guidelines and current issues in cancer screening. *CA Cancer J Clin* **66**, 96–114 (2016).
4. Patel, S.G., *et al.* Updates on Age to Start and Stop Colorectal Cancer Screening: Recommendations From the U.S. Multi-Society Task Force on Colorectal Cancer. *Am J Gastroenterol* **117**, 57–69 (2022).
5. Parekh, A., *et al.* The 50-Year Journey of Lung Cancer Screening: A Narrative Review. *Cureus* **14**, e29381 (2022).
6. Siegel, R.L., Kratzer, T.B., Giaquinto, A.N., Sung, H. & Jemal, A. Cancer statistics, 2025. *CA Cancer J Clin* **75**, 10–45 (2025).
7. Hoffman, R.M., *et al.* Multicancer early detection testing: Guidance for primary care discussions with patients. *Cancer* **131**, e35823 (2025).
8. Jung, A.W., *et al.* Multi-cancer risk stratification based on national health data: a retrospective modelling and validation study. *Lancet Digit Health* **6**, e396–e406 (2024).
9. Huguet, N., *et al.* Using Electronic Health Records in Longitudinal Studies: Estimating Patient Attrition. *Med Care* **58 Suppl 6 Suppl 1**, S46–S52 (2020).
10. Ye, J., Woods, D., Jordan, N. & Starren, J. The role of artificial intelligence for the application of integrating electronic health records and patient-generated data in clinical decision support. *AMIA Jt Summits Transl Sci Proc* **2024**, 459–467 (2024).
11. Rose, C. & Chen, J.H. Learning from the EHR to implement AI in healthcare. *NPJ Digit Med* **7**, 330 (2024).
12. Yang, X., *et al.* A large language model for electronic health records. *NPJ Digit Med* **5**, 194 (2022).
13. Guo, L.L., *et al.* EHR foundation models improve robustness in the presence of temporal distribution shift. *Sci Rep* **13**, 3767 (2023).
14. Renc, P., *et al.* Foundation model of electronic medical records for adaptive risk estimation. *Gigascience* **14**(2025).
15. Renc, P., *et al.* Zero shot health trajectory prediction using transformer. *NPJ Digit Med* **7**, 256 (2024).
16. Wornow, M., Thapa, R., Steinberg, E., Fries, J.A. & Shah, N.H. EHRSHOT: An EHR Benchmark for Few-Shot Evaluation of Foundation Models. *Adv Neur In* (2023).
17. Wornow, M., *et al.* The shaky foundations of large language models and foundation models for electronic health records. *NPJ Digit Med* **6**, 135 (2023).
18. Zhang, B., Shi, H. & Wang, H. Machine Learning and AI in Cancer Prognosis, Prediction, and Treatment Selection: A Critical Approach. *J Multidiscip Healthc* **16**, 1779–1791 (2023).
19. Goldstein, B.A., Navar, A.M., Pencina, M.J. & Ioannidis, J.P. Opportunities and challenges in developing risk prediction models with electronic health records data: a systematic review. *J Am Med Inform Assoc* **24**, 198–208 (2017).

20. Mishra, A.K., Chong, B., Arunachalam, S.P., Oberg, A.L. & Majumder, S. Machine Learning Models for Pancreatic Cancer Risk Prediction Using Electronic Health Record Data-A Systematic Review and Assessment. *Am J Gastroenterol* **119**, 1466–1482 (2024).
21. Canto, M.I.*, et al.* International Cancer of the Pancreas Screening (CAPS) Consortium summit on the management of patients with increased risk for familial pancreatic cancer. *Gut* **62**, 339–347 (2013).
22. Smith, R.A.*, et al.* Cancer screening in the United States, 2017: A review of current American Cancer Society guidelines and current issues in cancer screening. *CA Cancer J Clin* **67**, 100–121 (2017).
23. Ramirez, A.H.*, et al.* The All of Us Research Program: data quality, utility, and diversity. *Patterns* **3**(2022).
24. Seppala, T.T., Burkhart, R.A. & Katona, B.W. Hereditary colorectal, gastric, and pancreatic cancer: comprehensive review. *BJS Open* **7**(2023).
25. Park, J.*, et al.* Impact of population screening for Lynch syndrome insights from the All of Us data. *Nat Commun* **16**, 523 (2025).
26. Klann, J.G., Joss, M.A., Embree, K. & Murphy, S.N. Data model harmonization for the All Of Us Research Program: Transforming i2b2 data into the OMOP common data model. *PloS one* **14**, e0212463 (2019).
27. Wang, L.*, et al.* A scoping review of OMOP CDM adoption for cancer research using real world data. *NPJ Digit Med* **8**, 189 (2025).
28. Elnashar, A., White, J. & Schmidt, D.C. Enhancing structured data generation with GPT-4o evaluating prompt efficiency across prompt styles. *Front Artif Intell* **8**, 1558938 (2025).
29. Chen, T. & Guestrin, C. Xgboost: A scalable tree boosting system. in *Proceedings of the 22nd acm sigkdd international conference on knowledge discovery and data mining* 785–794 (2016).
30. Lau, K., Hart, G.R. & Deng, J. Predicting time-to-first cancer diagnosis across multiple cancer types. *Sci Rep* **15**, 24790 (2025).
31. Pang-Ning, T., Steinbach, M. & Kumar, V. Introduction to data mining: Pearson Addison Wesley. *Boston* (2006).
32. Patel, V.R., Adamson, A.S. & Welch, H.G. The Rise in Early-Onset Cancer in the US Population-More Apparent Than Real. *JAMA Intern Med* **185**, 1370–1374 (2025).
33. Pang, C.*, et al.* CEHR-XGPT: A Scalable Multi-Task Foundation Model for Electronic Health Records. *arXiv preprint arXiv:2509.03643* (2025).
34. Steinberg, E., Fries, J., Xu, Y. & Shah, N. MOTOR: A time-to-event foundation model for structured medical records. *arXiv preprint arXiv:2301.03150* (2023).
35. Pang, C.*, et al.* FoMoH: A clinically meaningful foundation model evaluation for structured electronic health records. *arXiv preprint arXiv:2505.16941* (2025).
36. Tomihari, A. & Sato, I. Understanding linear probing then fine-tuning language models from ntk perspective. *Advances in Neural Information Processing Systems* **37**, 139786–139822 (2024).
37. Mosca, E., Szigeti, F., Tragianni, S., Gallagher, D. & Groh, G. SHAP-based explanation methods: a review for NLP interpretability. in *Proceedings of the 29th international conference on computational linguistics* 4593–4603 (2022).
38. Carrington, J.M. & Effken, J.A. Strengths and limitations of the electronic health record for documenting clinical events. *Comput Inform Nurs* **29**, 360–367 (2011).

39. Kim, M.K., Rouphael, C., McMichael, J., Welch, N. & Dasarathy, S. Challenges in and Opportunities for Electronic Health Record-Based Data Analysis and Interpretation. *Gut Liver* **18**, 201–208 (2024).
40. Shobeiri, S. Enhancing Transparency in Healthcare Machine Learning Models Using Shap and Deeplift a Methodological Approach. *Iraqi Journal of Information and Communication Technology* **7**, 56–72 (2024).
41. Ozanne, E.M.*, et al.* Bias in the reporting of family history: implications for clinical care. *J Genet Couns* **21**, 547–556 (2012).
42. Park, J.*, et al.* Enhancing EHR-based pancreatic cancer prediction with LLM-derived embeddings. *NPJ Digit Med* **8**, 465 (2025).
43. Placido, D.*, et al.* A deep learning algorithm to predict risk of pancreatic cancer from disease trajectories. *Nat Med* **29**, 1113–1122 (2023).
44. Reinecke, I., Zoch, M., Reich, C., Sedlmayr, M. & Bathelt, F. The usage of OHDSI OMOP–a scoping review. *German Medical Data Sciences 2021: Digital Medicine: Recognize–Understand–Heal*, 95–103 (2021).